\UseRawInputEncoding
\documentclass[conference]{IEEEtran}
\usepackage{amsfonts}
\IEEEoverridecommandlockouts

\ifCLASSINFOpdf
\else
\fi

\usepackage{epsfig}
\usepackage{graphicx}
\usepackage{psfig}
\usepackage{epsf}
\usepackage[cmex10]{amsmath}
\usepackage{booktabs}
\usepackage{fancyhdr}
\usepackage{slashbox}
\usepackage{caption2}
\begin{document}
\title{Cognitive Semantic Communication Systems Driven by Knowledge Graph}
\author{Fuhui Zhou$^{\ddag}$, Yihao Li$^{\ddag}$, Xinyuan Zhang$^\dagger$, Qihui Wu$^{\ddag}$, Xianfu Lei$^\dagger$, and Rose Qingyang Hu$^{\S} $\\
$^{\ddag}$Nanjing University of Aeronautics and Astronautics, China,\\$^{\S} $Utah State University, Logan, UT 84322 USA\\$^\dagger$School of Information Science and Technology, Southwest Jiaotong University, Chengdu, China\\

Email: \emph{\{zhoufuhui@ieee.org, liyihao1999@nuaa.edu.cn, zxyleh@gmail.com,}
\\
\emph{wuqihui2014@sina.com, xflei@swjtu.edu.cn, rosehu@ieee.org \}}
\thanks{This work was supported by National Key Research and Development Program of China under Grant 2020YFB1807602, the National Natural Science Foundation of China under Grant 62031012, and  62071223 and Grant 61931011, Young Elite Scientist Sponsorship Program by CAST.}}
\maketitle
\begin{abstract}
Semantic communication is envisioned as a promising technique to break through the Shannon limit. However, the existing semantic communication frameworks do not involve inference and error correction, which limits the achievable performance. In this paper, in order to tackle this issue, a cognitive semantic communication framework is proposed by exploiting knowledge graph. Moreover, a simple, general and interpretable solution for semantic information detection is developed by exploiting triples as semantic symbols. It also allows the receiver to correct errors occurring at the symbolic level. Furthermore, the pre-trained model is fine-tuned to recover semantic information, which overcomes the drawback that a fixed bit length coding is used to encode sentences of different lengths. Simulation results on the public WebNLG corpus show that our proposed system is superior to other benchmark systems in terms of the data compression rate and the reliability of communication.
\end{abstract}
\IEEEpeerreviewmaketitle%生成单独封面
\section{Introduction}
The wireless communication system has been flourishing and evolved from the first generation (1G) to the latest fifth generation \cite{acticle21}. Those systems has been developed based on the Shannon information theory. With the development of the advanced channel coding scheme, such as low-density parity-check (LDPC) codes and polar codes, the conventional communication system has gradually approached the Shannon limit \cite{acticle3}. It is imperative to make breakthrough for this limit to meet the unprecedented proliferation of mobile devices, the increasing requirements of high data rate and the emerging of diverse ultra-band wideband services \cite{acticle4}.

Semantic communication proposed by Weaver and Shannon is promising to tackle this issue and has been receiving ever increasing attention from both academic and industry \cite{acticle16}. Different from the conventional wireless communication systems that focus on the successful transmission of symbols from the transmitter to the receiver, the semantic communication system pays more attention to the semantic information at the transmitter and the meaning interpreted at the receiver \cite{acticle4}. It can significantly reduce the amount of data required to be transmitted and also improve the performance robustness.

Up to known, semantic communications can be realized by three paradigms. The first one is the goal-oriented communication paradigm that only transmits the important and necessary messages based on the goal-oriented metrics, such as the control-theoretic accuracy proposed in \cite{acticle5} and the age-of-information-based metrics proposed in \cite{acticle6}, \cite{acticle7}. In this paradigm, semantic compression is achieved by filtering out redundant messages that are less relevant to the communication goal. The second paradigm is to embed raw data into a low-dimensional space by using deep neural networks in order to compress source information and achieve semantic communication. The authors in \cite{acticle8}-\cite{acticle18} proposed a joint source channel codec framework based on the natural language processing (NLP) technology, such as bi-directional long short-term memory (BiLSTM) and transformer. The last one is to exploit a sharing knowledge base between the source and destination to achieve semantic compression and communication \cite{acticle4}.

The first paradigm achieves semantic compression explainably and communication based on the goal-oriented metrics. However, there have been no uniform metrics for evaluating the achievable performance. Due to this fact, the first paradigm is restricted to some specific wireless services or communication environment \cite{acticle10}. The second paradigm relys on the large-scale deep learning (DL) models to accurately and quickly recognize and extract intended semantic information, and it does not require semantic metrics. Nevertheless, DL normally requires a large number of high quality labeled data. Moreover, the end-to-end operation is identified as a black-box process, which lacks interpretability. The last one achieves semantic compression and communication with the help of knowledge base without semantic metrics. Moreover, it has the interpretability during the compression process due to the reasoning rules in the knowledge base \cite{acticle4}. The third paradigm is envisioned to be the most promising one \cite{acticle10}. However, there are few investigations that have focused on this direction since it requires the interdisciplinary knowledge, such as wireless communication, computer science, artificial intelligence, etc.

\begin{figure}[htbp]
\centering
\includegraphics[width=3.6 in]{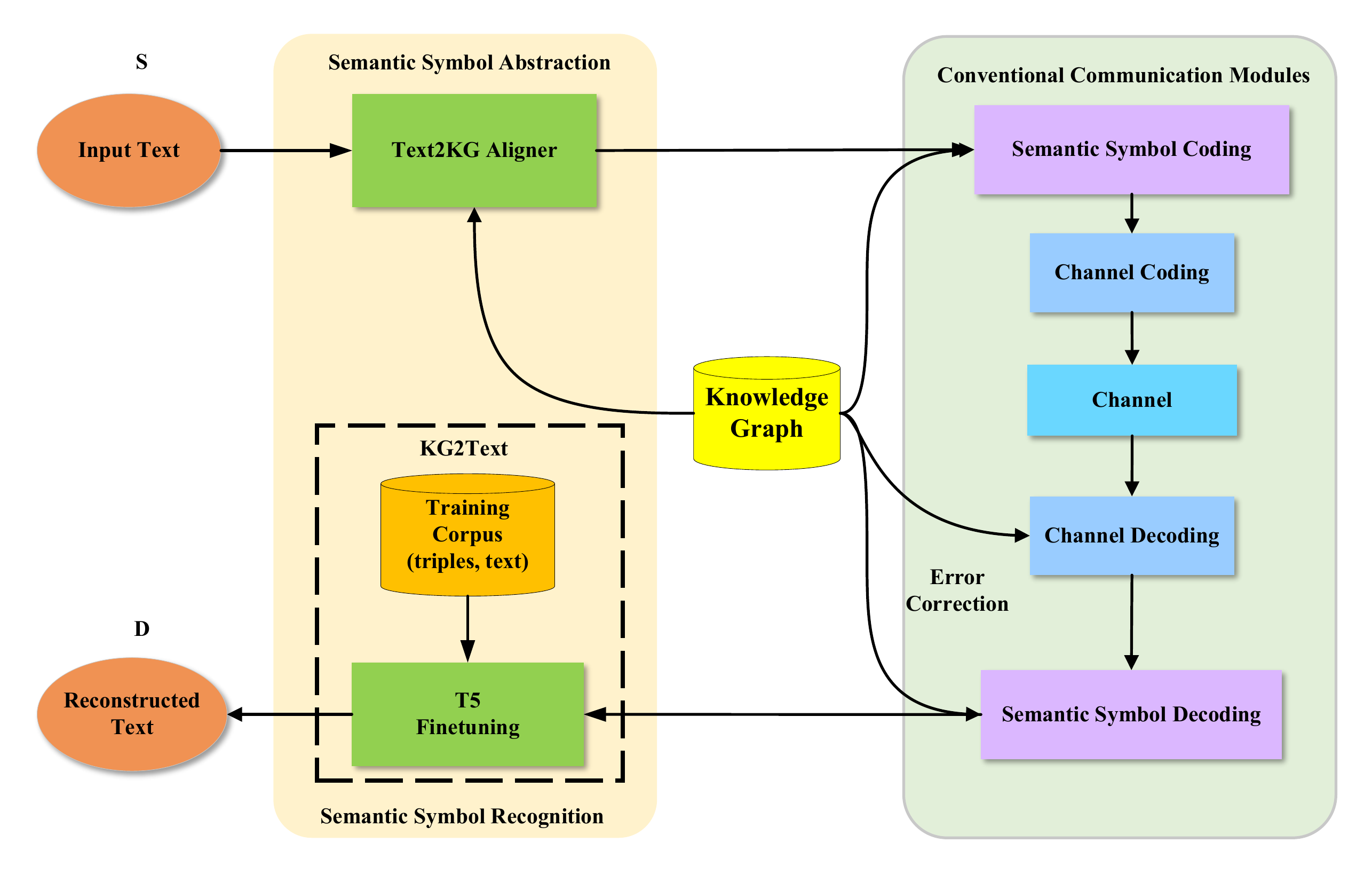}
\caption{The proposed cognitive semantic communication framework driven by the knowledge graph.} \label{fig.1}
\end{figure}

Motivated by those facts, a cognitive semantic communication system is proposed by using knowledge graph, which is a sharing knowledge base between the transmitter and the receiver. In order to develop a simple and general solution for semantic information detection, the triples are viewed as semantic symbols. Since the triples are more general form of semantic organization and are inherently readable, our proposed system is interpretable. Moreover, the pre-trained model is fine-tuned to recover semantic information, which overcomes the drawback that a fixed bit length coding is used to encode sentences of different lengths. Simulation results demonstrate that our proposed system is superior to other benchmark systems in terms of the data compression rate and the reliability of communication.

The remainder of this paper is organized as follows. Our proposed cognitive semantic communication system is presented in Section II. Section III presents knowledge graph and system implementation. Section IV presents simulation results. Finally, the paper concludes with Section V.

\section{Cognitive Semantic Communication}
In this section, a cognitive semantic communication system is proposed based on the sharing knowledge graph between the transmitter and the receiver. The framework for this system is shown in Fig. 1.

\subsection{Cognitive Semantic Communication Framework}
%认知语义通信的定义
In the traditional semantic communication system, the transmitter sends the abstraction of semantic information and the receiver interprets its meaning without inference. Our proposed cognitive semantic communication system has the \lq\lq cognitive\rq\rq \ characteristic, which is enabled by knowledge graph. In the proposed system, the information is not required to be completely transmitted and only the important semantic information (such as the triple, namely, the head entity, relation and tail entity) is transmitted. The receiver interprets the meaning by inferring from the received information. Due to the inference characteristic, our proposed cognitive semantic communication system  has three evident advantages. Firstly, a high semantic compression rate can be achieved since only the important information is transmitted. Secondly, a robust performance can be obtained due to the reason that the semantic error can be corrected via inference driven by knowledge graph. Finally, the interpretability of the communication process can be realized since the triples of the knowledge graph have the comprehensibility. The details for the proposed semantic communication system are presented as follows.

The framework of the proposed cognitive semantic communication system is shown in Fig. 1. A source $S$ generates a message $m\in M$. $M$ is the set of all the messages. $m$ carries the semantic information that $S$ transmits to $D$. In this paper, the text transmission is considered as an example. In order to transmit the semantic of message, semantic requires first to be expressed as a formal way. According to the work in \cite{acticle3}, semantics (meaning) are associated to a knowledge system. Thus, in our work, the knowledge graph is exploited as the knowledge system and the triples are exploited to represent semantic. The semantic symbol is denoted as $s$, and $s$ is included in $S$, where $S$ is a set of the semantic symbols. In order to improve effectiveness of communication, the semantic symbol is abstracted from the message $m$. This process is denoted as $s=f\left( m \right)$. The Text2KG aligner is designed to implement the abstraction of the semantic symbol $s$ from message $m$.

After the semantic symbol is obtained, the semantic symbol is transmitted by using the conventional communication modules (CCMs). Specifically, the semantic symbol $s$ is encoded as $x$ in order to improve the transmission efficiency. Then, the channel coding is performed and $b$ is obtained. At the destination, the binary vectors are received and the semantic symbol code is obtained by channel decoding. Note that the knowledge graph is exploited to correct error in this process. After that, the semantic symbol is obtained by using semantic symbol decoding.

\begin{figure}[htbp]
\centering
\includegraphics[width=2.4 in]{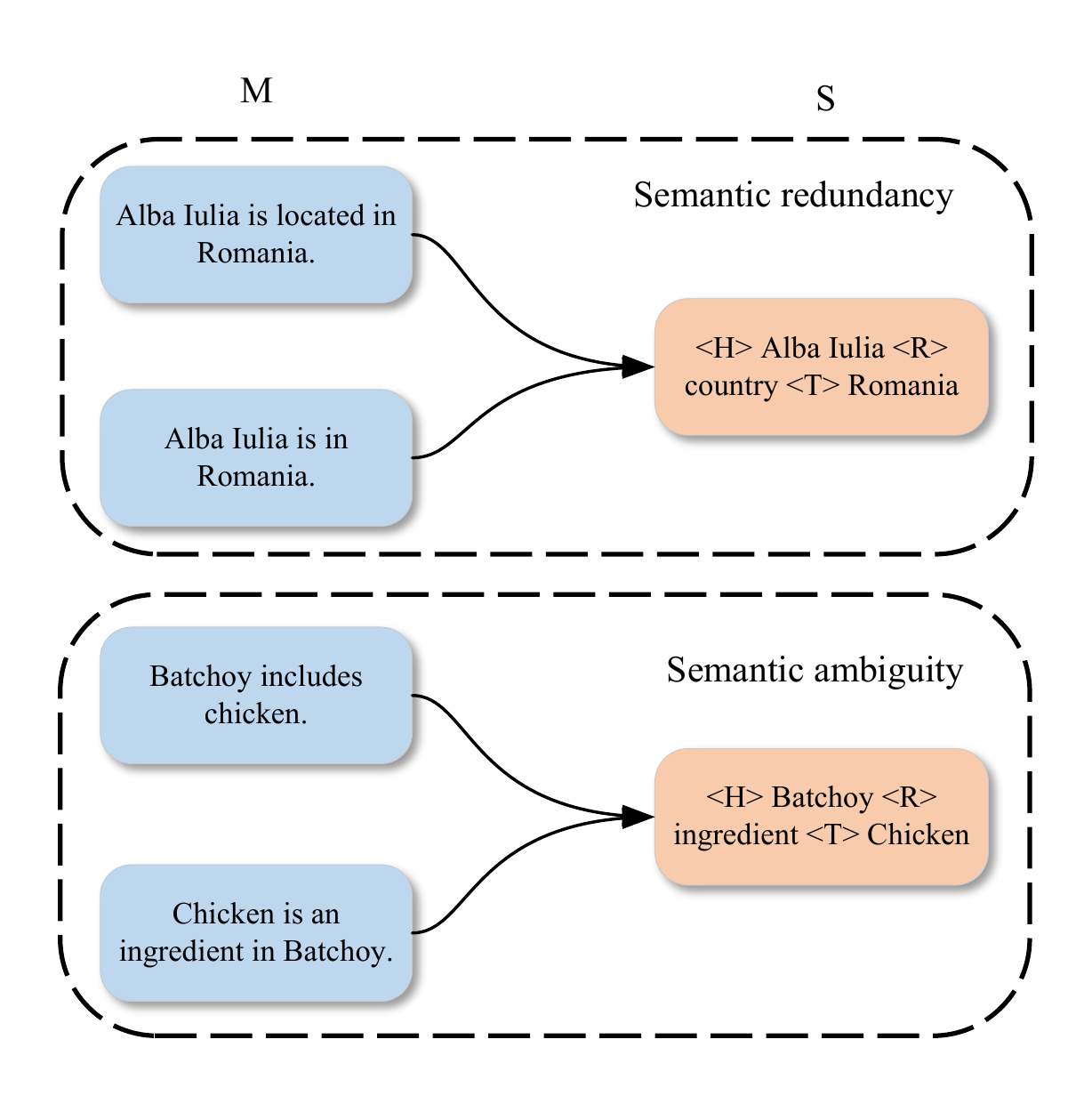}
\caption{An example diagrams the mapping message $m\in M$ to semantic symbol $s\in S$}
\label{fig.0}
\end{figure}

Finally, in order to reconstruct message $\hat{m}\in \hat{M}$ from the obtained semantic symbol, where $\hat{M}$ is the reconstructed message set, the semantic symbol recognition module is designed based on the NLP technology. In this paper, different from the work in \cite{acticle9} that the joint source-channel codec framework is achieved by end-by-end training transformer, a novel framework by training a NLP technology at the receiver is proposed. Our framework can overcome the drawback that a fixed bit length coding is used to encode sentences of different lengths and results in low efficiency.

Note that since the mapping from $m$ to $s$ is many-to-one, the semantic ambiguity may exist. In order to mitigate semantic ambiguity and implement the triple-to-text conversion, the pre-training model Text-to-Text Transfer Transformer model (T5) is fine-tuned on our training corpus \cite{acticle13}. Since the pre-training model T5 is fed by billions of sentences, it can take context into account when the reconstructed text is generated. For example, as shown in Fig. 2, $S$ generates the message \lq Batchoy includes chicken\rq. \ However, \lq chicken\rq \ can be translated as animal or meat in different contexts. It results in the semantic ambiguity. The fine-tuned T5 model is used to reconstruct the text \lq Chicken is an ingredient of Batchoy.\rq \ at the receiver side. It is observed that the semantic ambiguity can be removed. In contrast, the conventional communication system cannot effectively tackle the semantic redundancy or semantic ambiguity.

\subsection{The Feasibility and Reasonability of Our Proposed Framework.}
The semantic information theory is used to verity the feasibility and reasonability of our proposed cognitive semantic communication system. It is assumed that the source randomly transmits a message $m$ with the probability $\Pr(m)$. Thus, a massage entropy $H\left( M \right)$ of the source can be expressed as
$$H\left( M \right)=-\underset{m\epsilon M}{\mathop \sum }\,\Pr(m)\log_2\Pr(m).\eqno(1)$$
According to the work in \cite{acticle3}, the logical probability $\Pr(s)$ of the semantic symbol $s$ can be expressed as
$$\Pr(s)=\underset{s=f\left( m \right),m\in M}{\mathop \sum }\,\Pr(m).\eqno(2)$$
Thus, based on eq. (2), the semantic entropy $H\left( S \right)$ related to the semantic symbol set $S$ is
$$H\left( S \right)=-\underset{s\epsilon S}{\mathop \sum }\,\Pr(s)\log_2\Pr(s).\eqno(3)$$
According to the classic information theory, $H\left( S \right)$ can be written as
$$H\left( S \right)=H\left( S/M \right)+I\left( M;S \right),\eqno(4)$$
where $H\left( S/M \right)$ represents the uncertainty of the random variable $S$ under the condition of the known random variable $M$ and $I\left( M;S \right)$ represents the average mutual information between $S$ and $M$. Since $I\left( M;S \right)=H\left( M \right)-H\left( M/S \right)$, the semantic entropy can be rewritten as
$$H\left( S \right)=H\left( M \right)+H\left( S/M \right)-H\left( M/S \right).\eqno(5)$$
Since the mapping from $m$ to $s$ is often many-to-one by semantic symbol abstraction, $H\left( S \right)$ is always smaller than $H\left( M \right)$. Thus, there exists an entropy loss $H\left( M \right)-H\left( S \right)$. The reducing entropy of the source is a desired compression of the source without a \lq real\rq \ semantic loss, since the messages that contain the same semantics have many equivalent forms as shown in Fig. 2. For example, \lq Alba Iulia is located in Romania.\rq \ and \lq Alba Iulia is in Romania.\rq \ are different in the form. However, they have the same semantic information. Our proposed Text2KG aligner can align them to the same triple \lq $<$H$>$ Alba Iulia $<$R$>$ country $<$T$>$ Romania \rq.

\begin{figure}[htbp]
\centering
\includegraphics[width=3.6 in]{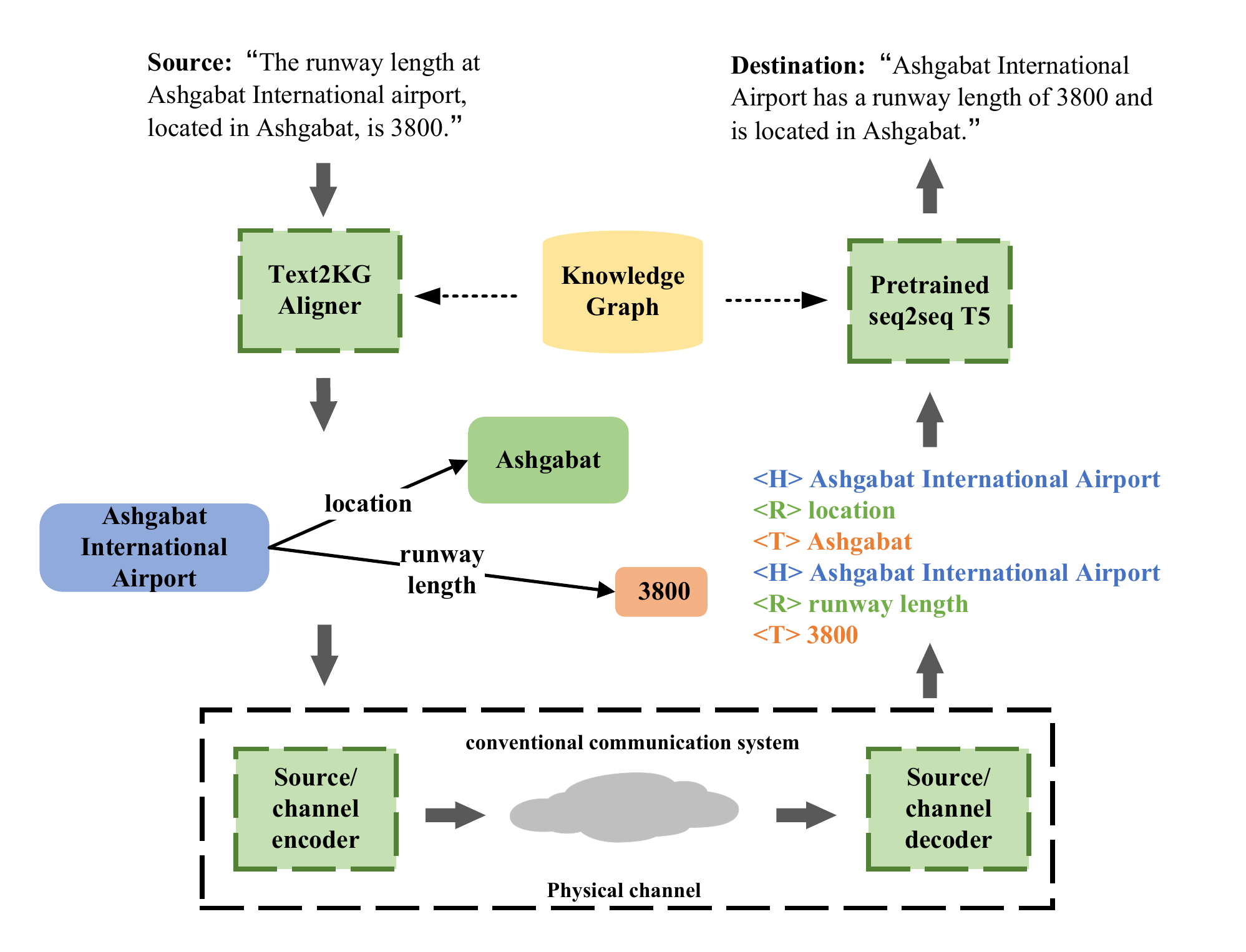}
\caption{An example for text transmission achieved by using our proposed cognitive semantic communication system.}
\label{fig.2}
\end{figure}

\section{Knowledge Graph and System Implementation}
In this section, the details for the knowledge graph and the implementation of our proposed cognitive semantic communication system are presented. An example for text transmission achieved by using our proposed cognitive semantic communication system is shown in Fig. 3.

\subsection{Knowledge Graph}
Knowledge graph, as the core of cognitive semantic communication, is a semantic network that reveals the relationship among entities in the form of graphs. It generally consists of a triple (head, relation, tail) or (entity, attribute, value) to express facts. Its construction methods are usually divided into top-down and bottom-up approaches. Large-scale general knowledge graph used in our cognitive semantic communication is always constructed top-down including four procedures, namely, information extraction, knowledge fusion, knowledge processing, and knowledge update \cite{acticle17}.

\subsection{Semantic Symbol Abstraction}
The semantic symbol abstraction aims to detect and extract the specific semantic information contained in the message. The implementation of the semantic symbol abstraction is to align the input text $m$ with the triplet$(h,r,t)$, where $h$, $r$ and $t$ denote the head entity, the relation and the tail entity of the knowledge graph, respectively. It is realized by an Text2KG alignment algorithm, as shown in Table 1 \cite{acticle13}. Note that for each sentence in the input text, all triples that have $h$ and $t$ are matched. Moreover, each triple can align to multiple sentences and each sentence can have multiple triples aligned to it. The relations are not required to be matched since there are many ways to express them. Similar to the work in \cite{acticle13}, the relation can be expressed as long as the message mentions the head entity and tail entity.
\begin{table}[htbp]
\begin{center}
\caption{The alignment algorithm}
\begin{tabular}{lcl}
\\\toprule
$\textbf{Algorithm 1}$: The Text2KG alignment algorithm  \ \ \ \ \ \ \ \ \ \ \ \ \ \ \ \ \ \ \ \ \ \\ \midrule
alignment\_triples $\gets$ $\left[ \ \right]$ \\
\textbf{for all} sentences $s$ $\in$ input text \textbf{do}\\
\ \ \ \ \ \textbf{for all} triples$(h,r,t)$ $\in$ KG \textbf{do}\\
\ \ \ \ \ \ \ \ \ \ \textbf{if} s.contains($h$) \textbf{and} s.contains($t$) \textbf{then}\\
\ \ \ \ \ \ \ \ \ \ \ \ \ \ \ alignment\_triples.add($(h,r,t)$)\\
\ \ \ \ \ \ \ \ \ \ \textbf{end if}\\
\ \ \ \ \ \textbf{end for}\\
\textbf{end for}\\
\bottomrule
\end{tabular}
\end{center}
\end{table}
\subsection{Conventional Communication Modules}
After the semantic symbols are obtained, in order to complete semantic communications, those symbols are transmitted by using the conventional communication modules, namely, semantic symbol coding, channel coding, channel, channel decoding and semantic symbols decoding. Specifically, according to \cite{acticle1}, for a propositional logic with finite propositions, the size of all possible interpretations (worlds) is finite. Thus, the number of entities and relations in the knowledge graph which describes the world is finite and it ensures that the semantic symbol code has a limited length. Firstly, in order to achieve semantic symbol coding, a dictionary $\left( key, value \right)$ is created. It enables the head entity, tail entity and relation of the knowledge graph to be uniquely mapped as integers. Secondly, a binary vector $x$ is obtained by encoding each integer with a fixed length. This process is denoted as the semantic symbol encoding (SSC).

In order to improve the communication reliability, the channel coding module is applied and the coded binary vector set $B$ is obtained. Then, the set $B$ is transmitted over the channel. Note that the binary channel is considered in this paper. Since the channel is noisy, errors can occur during transmission. In order to improve the robustness of our proposed system, channel decoding is implemented by exploiting knowledge graph
to correct error. The process of the correction is to traverse the semantic symbols in the knowledge graph to find the most similar one observed at the receiver. The details for the correction algorithm denoted as Algorithm 2 can be found in Table II. Semantic symbol decoding is the reverse process of semantic symbol encoding. After semantic symbol decoding, the reconstructed semantic symbol $\hat{S}$ is achieved.

\begin{table}[htbp]
\begin{center}
\caption{The correction algorithm}
\begin{tabular}{lcl}
\\\toprule
$\textbf{Algorithm 2}$: The correction algorithm based on the KG  \ \ \ \ \ \ \ \ \ \ \ \ \ \ \\ \midrule
$\textbf{Input}$: $o$ is the binary vector observed at the receiver, \\ $KG\_coding$ is the coding of all triples in the KG \\
$\textbf{function}$ similar ($a$, $b$)\\
\ \ \ \ \ sim $\gets$ 0 \\
\ \ \ \ \ \ \ \ \ \ $\textbf{for}$ $i$ $\gets$ 0 $\textbf{to}$ len($b$) $\textbf{by}$ 1 $\textbf{do}$ \\
\ \ \ \ \ \ \ \ \ \ $\textbf{if}$ a[$i$] != b[$i$] $\textbf{then}$ \\
\ \ \ \ \ \ \ \ \ \ \ \ \ \ \ sim++ \\
\ \ \ \ \ $\textbf{end for}$ \\
\ \ \ \ \ $\textbf{return}$ sim \\
$\textbf{end function}$ \\
tmp $\gets$ $\left[ \ \right]$ \\
$\textbf{for}$ each $code$ $\in$ $KG\_coding$ $\textbf{do}$ \\
\ \ \ \ \ tmp.add(similar ($o$, $code$))\\
$o$ $\gets$ $KG\_encode$[tmp.index(max(tmp))]\\
$\textbf{Output}$: $o$ is the corrected binary vector \\
\bottomrule
\end{tabular}
\end{center}
\end{table}
\subsection{Semantic Symbol Recognition}
After the semantic symbol is obtained, in order to interpret the meaning of semantic information, the crucial process of our proposed system is to transform triples into text. This process needs to exploit the NLP technology. In this paper, the T5 model \cite{acticle14} is fine-tuned on the local dataset in order to convert triples in the knowledge graph into natural text. Specifically, the triples in the dataset need to be concatenated as $head \ relation\_1 \ tail\_1,....,relation\_n \ tail\_n$ for inputing into T5.

\section{Simulation Results}
In this section, simulation results are presented to evaluate the performance of the proposed cognitive semantic communication system and compare it with that of the traditional communication system realized by the separate source and channel coding technologies under the binary symmetric channel (BSC).

\subsection{Dataset and Finetuning Parameters}
The training set of WebNLG English dataset \cite{acticle13} is used to fine-tune the T5 model. It contains knowledge graphs and text from various domains including Airport, Artist, Astronaut, Athlete, Building, Celestial Body, City, Comics Character, Food, Mode of Transportation, Monument, Politician, Sports Team, University and Written Work \cite{acticle13}. The test sets include three additional domains, namely, Film, Scientist, and Musical Work. The T5 model is fine-tuned on the training sets for 100 steps with a learning rate of 0.0001. We compare the performance of the proposed system with the benchmark systems by transmitting the text in the test set.
\begin{table}[htbp]
\begin{center}
\begin{tabular}{lcl}
\\\toprule
$\textbf{Split}$\ \ \ \ \ \ \ \ \ \ \ \ \ \ \ \ \ \ \ \ \ \ \ \ \ \ \ \ \ \ \ \ \ \ \ \ \ \ \ \ \ \ \ $\textbf{Number of Samples}$\\ \midrule
Train(T-G Pairs)\ \ \ \ \ \ \ \ \ \ \ \ \ \ \ \ \ \ \ \ \ \ \ \ \ \ \ \ \ \ \ \ \ \ \ \ \ \ \ 35426\\
Test(T-G Pairs)\ \ \ \ \ \ \ \ \ \ \ \ \ \ \ \ \ \ \ \ \ \ \ \ \ \ \ \ \ \ \ \ \ \ \ \ \ \ \ \ \ 4464\\
\bottomrule
\end{tabular}
\end{center}
\caption{Number of texts and knowledge graph pairs in the training, test sets of the WebNLG English dataset.}
\end{table}
\subsection{Performance Comparison}
In order to guarantee the fairness of performance comparison, the channel coding scheme of our proposed system is the same as that of the benchmark communication systems, namely, the binary convolutional codes under BSC. Moreover, two source coding schemes are exploited for the benchmark systems, namely, Huffman and fixed-length coding. In this paper, the fixed-length coding scheme adopts 7 bits for each char since the corpus is case-sensitive and contains special characters character. It is well known that the variable-length coding is more efficient than the fixed-length coding. To demonstrate the superiority, the
fixed-length coding is chosen as the source coding of our proposed system.

\begin{figure}[!t]
\centering
\includegraphics[width=2.8 in]{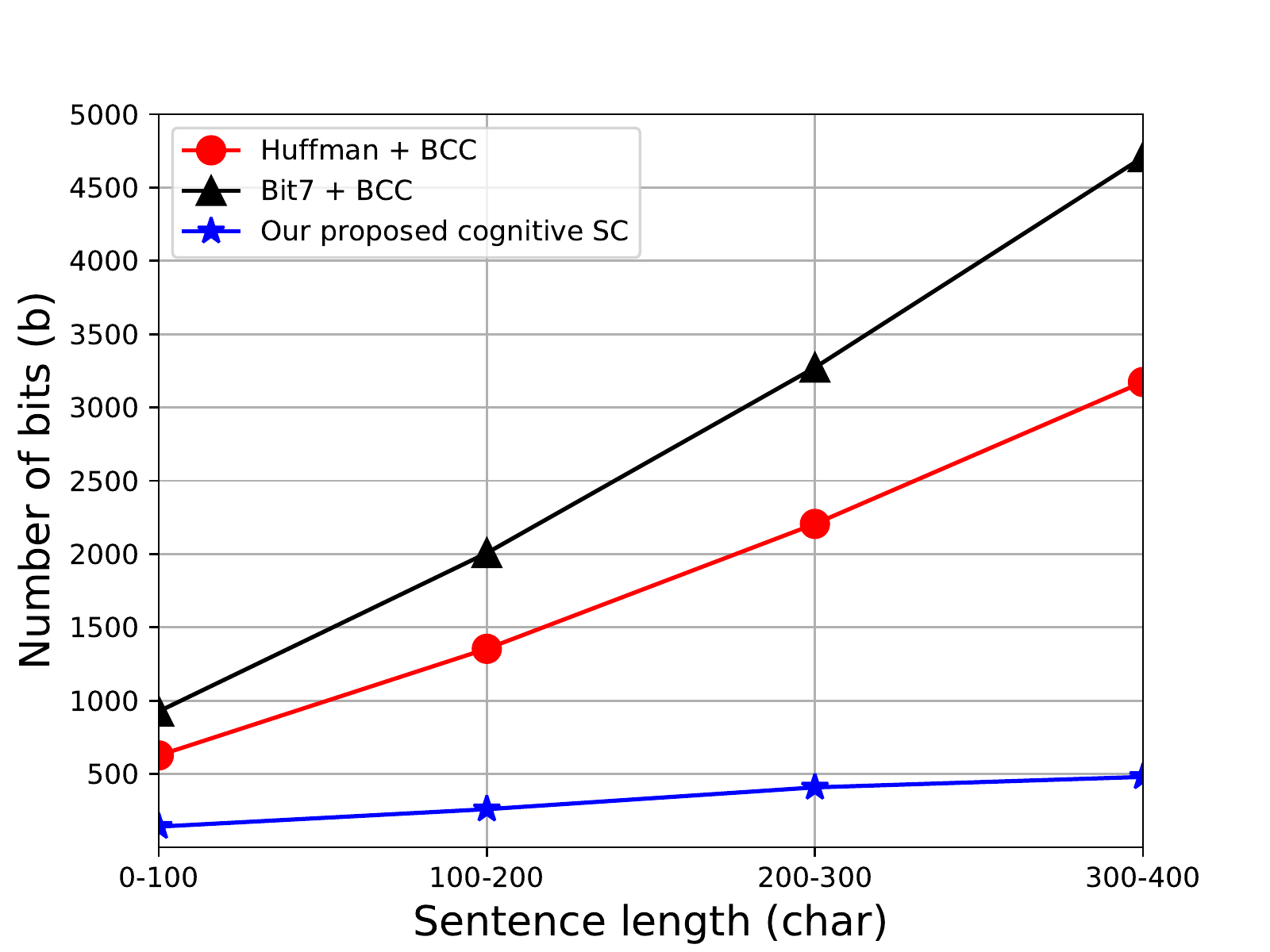}
\caption{The number of bits required to be transmitted by using different schemes versus the sentence length.} \label{fig.3}
\end{figure}

\begin{figure}[!t]
\centering
\includegraphics[width=2.8 in]{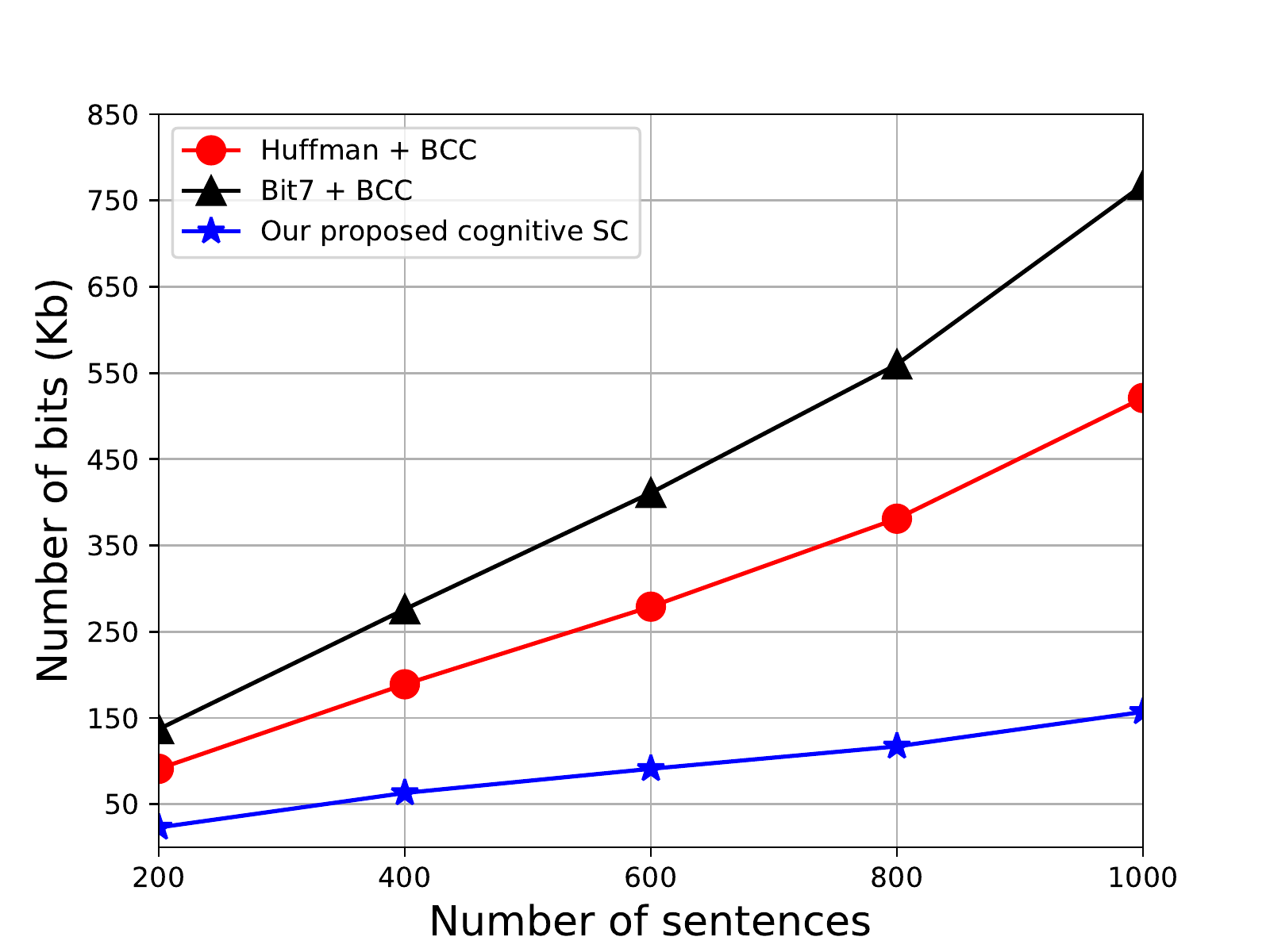}
\caption{The number of bits required to be transmitted by using different schemes versus the number of texts.} \label{fig.4}
\end{figure}

\begin{figure}[!t]
\centering
\includegraphics[width=2.8 in]{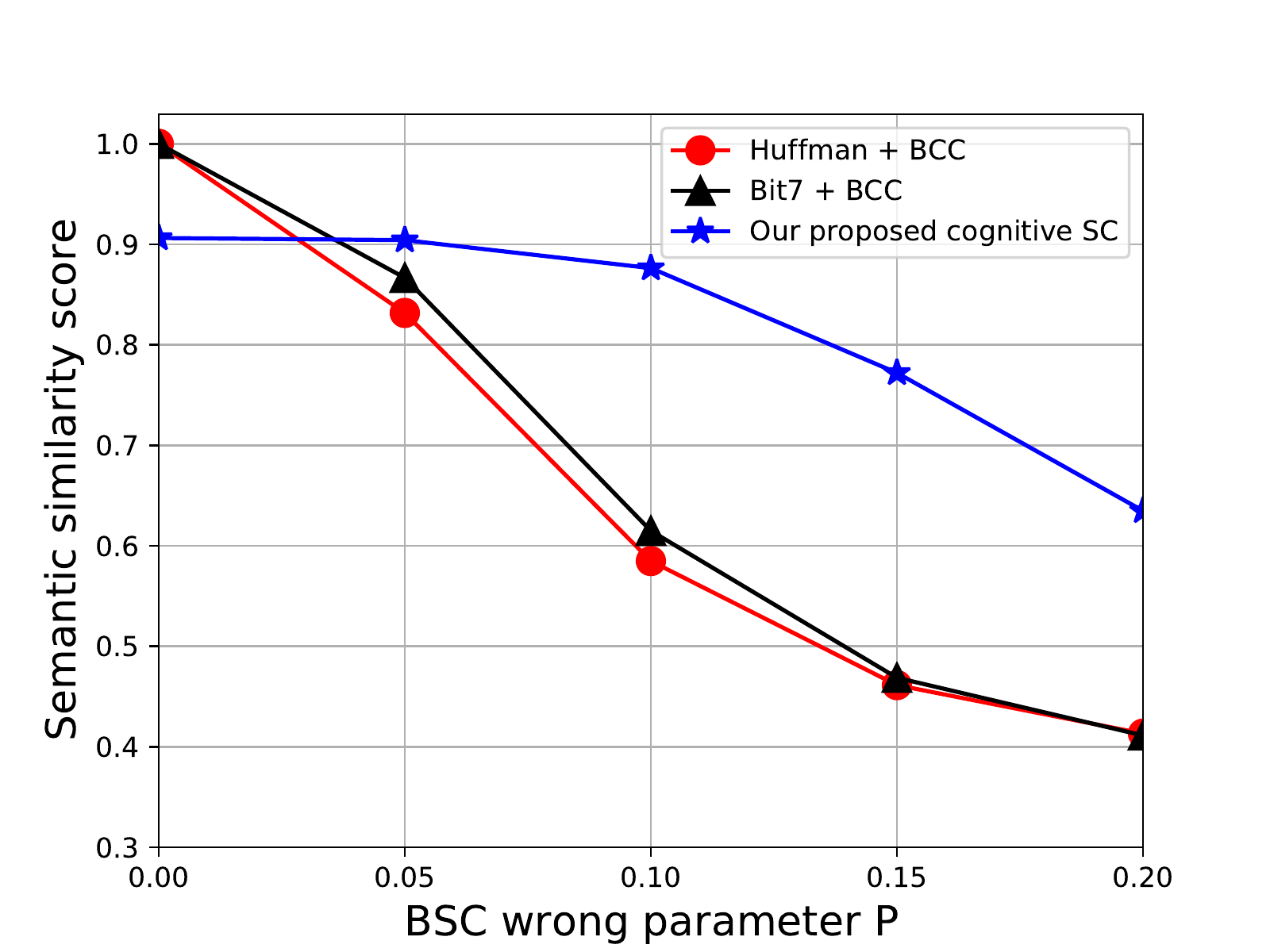}
\caption{The semantic similarity score versus the BSC wrong parameter $p$.} \label{fig.5}
\end{figure}

\begin{figure}[!t]
\centering
\includegraphics[width=3.8 in]{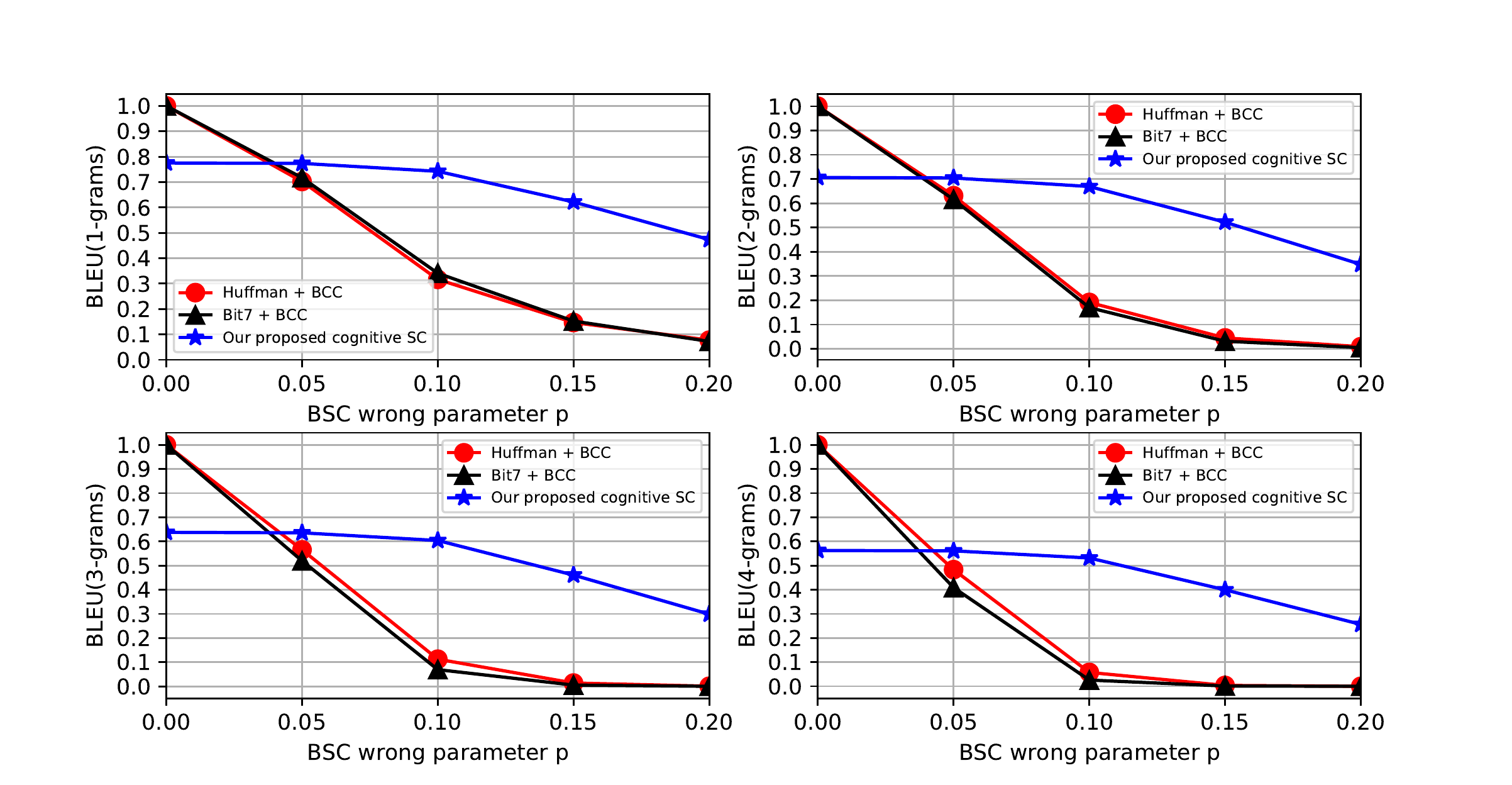}
\caption{The BLEU score versus the BSC wrong parameter $p$.} \label{fig.6}
\end{figure}

\begin{table*}\normalsize\centering
\begin{tabular}{|c|c|}
\hline 					% 横线
Lossless  & TX: The leader of Aarhus is Jacob Bundsgaard. \\ & RX: The leader of Aarhus is Jacob Bundsgaard.\\
\hline
Rephrasing but lossless semantics &   TX: Aarhus Airport's runway length is 2702.0.\\ & RX: The runway length of Aarhus Airport is 2702.0.\\
\hline
Align redundant information & TX: Adolfo Suarez Airport is found in San Sebastian de los Reyes.\\
& RX: Adolfo Suarez Airport is located in San Sebastian de los Reyes, Madrid.\\
\hline
An inexplicable error & TX: Aarhus Airport's runway length is 2702.0.\\ & RX: Asam pedas is a dish from Sumatra.\\
\hline
Disambiguation & TX: Batchoy includes chicken.\\ &RX: Chicken is an ingredient of Batchoy.\\
\hline
\end{tabular}
\caption{Cases which are transmitted and received using our proposed system when $p=0.1$.}
\end{table*}

In traditional communication systems, the symbol error rate (SER) is used to evaluate the communication performance. However, it is inappropriate for semantic communication since different sentences at the transmitter and receiver possibly have the same semantic information \cite{acticle9}. In order to tackle this issue, the Bilingual evaluation understudy (BLEU) and semantic similarity score are widely used to evaluate the performance of semantic communication and that of the traditional communication. BLEU score is usually used to evaluate the machine translation results. However, the BLEU score can only compare the difference among words of two sentences rather than their semantic information. Thus, in this paper, the semantic similarity score is used to measure performance \cite{acticle9}. According to the cosine similarity, the sentence similarity between the original sentence $m$ and the recovered sentence $\hat{m}$ is calculated as：
$$score\left( \hat{m},m \right)=\frac{{{B}_{\text{ }\!\!\Phi\!\!\text{ }}}\left( m \right){{B}_{\text{ }\!\!\Phi\!\!\text{ }}}{{\left( {\hat{m}} \right)}^{T}}}{\left| \left| {{B}_{\text{ }\!\!\Phi\!\!\text{ }}}\left( m \right)\left| \left| ~ \right| \right|{{B}_{\text{ }\!\!\Phi\!\!\text{ }}}\left( {\hat{m}} \right) \right| \right|},\eqno(6)$$
where ${{B}_{\text{ }\!\!\Phi\!\!\text{ }}}$, representing bidirectional encoder representations from transformers (BERT), is a large pre-trained model used for extracting the semantic information. It is trained on an ultra large-scale corpus. Compared with the BLEU score, BERT has been fed by billions of sentences \cite{acticle9}. Therefore, it can recognize semantic information effectively.

Fig. 4 shows the number of bits required to be transmitted versus the sentence length. The sentence length depends on the number of characters. It is quite evident that our proposed cognitive semantic communication system obtains the highest text compression rate among the benchmark systems, especially for long sentences. The reason is that the proposed system can capture the semantic equivalence and realize the semantic compression. Moreover, knowledge graphs used as the shared knowledge base on both the transmitter and receiver are benefical for further improving the coding efficiency. Fig. 5 shows the number of bits required to be transmitted versus the number of texts. The number of texts depends on the number of sentences. As shown in Fig. 5, the proposed cognitive semantic communication system transmits the same number of texts with less data compared with all benchmark systems. It further indicates that the proposed system has a superior compression rate.

Fig. 6 shows that the semantic similarity score versus the BSC wrong parameter $p$. The semantic similarity score is used to measure the semantic recovery performance. Note that $p$ represents the channel wrong parameter and a larger $p$ represents a worse the channel environment. As shown in Fig. 6, the scores of the benchmark systems are all close to 1 and higher than our proposed system when $p<0.05$. Since the channel coding corrects the errors that occur during transmission, the benchmark systems can recover the text without any loss at the receiver. However, the semantic information detection in our system has data loss which is actually the reason that our system can transmit less data with high semantic fidelity. Meanwhile, the score of our system is also higher than 0.9, which means that the messages can be well understood. When $p>0.05$, the channel environment is poor, the channel coding cannot correct all the error. In this case, the performance of the benchmark systems degrades rapidly. In contrast, since our proposed cognitive semantic communication system can more effectively capture and recover semantic information and confront errors, it can generate easy-to-understand messages. Fig. 7 shows the same tendency with BLEU score as the criteria for performance measurement. It is obviously seen that our proposed system is more competitive and robust in poor channel environments. A few cases between the transmitter and receiver are shown in Table IV in order to demonstrate the efficiency of our proposed cognitive semantic communication system.

\section{Conclusion}
A cognitive semantic communication system was proposed by using knowledge graph. Moreover, a simple, general and interpretable solution for semantic information detection was proposed to improve data compression rate. Furthermore, the T5 model was adjusted to recover semantic information in order to overcome the drawback that a fixed bit length coding is used to encode sentences of different lengths and results in low efficiency. The proposed cognitive semantic communication system underpins promising performance. Simulation results demonstrated that the proposed system has a superior data compression rate and communication reliablity.

\end{document}